\pdfoutput=1

\documentclass[11pt]{article}

\usepackage[preprint]{acl}

\usepackage{times}
\usepackage{latexsym}

\usepackage[T1]{fontenc}

\usepackage[utf8]{inputenc}

\usepackage{microtype}

\usepackage{inconsolata}

\usepackage{graphicx}

\usepackage{amsmath}
\usepackage{multirow}
\graphicspath{ {./imgs/} }

\newcommand{\lb}{\textcolor{black}}
\newcommand{\sara}{\textcolor{black}}

\newcommand{\mmg}{\textcolor{black}}

\title{Taming Convergence: \\ Learning Rate Warmup Strategies for Large-Scale Speech-to-Text Models}
\title{From Exploding Gradients to Smooth Convergence: \\ Rethinking Learning Rate Warmup for Large-Scale Speech-to-Text Models}
\title{The Role of Learning Rate Warmup in Large Scale Speech-to-text Training}
\title{The Warmup Dilemma: How Learning Rate Strategies Impact Speech-to-Text Model Convergence}

\author{Marco Gaido\textsuperscript{*}, Sara Papi\textsuperscript{*}, Luisa Bentivogli, Alessio Brutti, Mauro Cettolo, \\\textbf{Roberto Gretter, Marco Matassoni, Mohamed Nabih, Matteo Negri} \\
  Fondazione Bruno Kessler, Italy \\
  \texttt{\{mgaido,spapi,bentivo,brutti,cettolo,gretter,matasso,mnabih,negri\}@fbk.eu}}

\begin{document}
\maketitle
\begin{abstract}
Training large-scale models presents challenges not only in terms of resource requirements but also in terms of their convergence. For this reason, the learning rate (LR) is often decreased when the size of a model is increased. Such a simple solution is not enough in the case of speech-to-text (S2T) trainings, where evolved and more complex variants of the Transformer architecture -- e.g., Conformer or Branchformer -- are used in light of their better performance. As a workaround,
OWSM
designed a double linear warmup of the LR, increasing it to a very small value in the first phase before updating it to a higher value in the second phase. While this solution worked well in practice, it was not compared with alternative solutions, nor was the impact on the final performance of different LR warmup schedules studied. This paper fills this gap, 
revealing that \textit{i)} large-scale S2T trainings
demand a sub-exponential LR warmup, and 
\textit{ii)} a higher LR in the warmup phase accelerates initial convergence, but it does not boost final performance.

\end{abstract}

\newcommand\blfootnote[1]{%
  \begingroup
  \renewcommand\thefootnote{}\footnote{#1}%
  \addtocounter{footnote}{-1}%
  \endgroup
}
\blfootnote{* Equal contribution.}

\section{Introduction}

Following the success of Large Language Models (LLM) \citep{radford2019language}, large-scale speech-to-text (S2T) trainings have gained increased interest with the goal of building Large Speech Models (LSM) or Speech Foundation Models (SFM) with similar abilities for the speech modality \citep{communication2023seamlessm4t,peng2023owsm,whisper,zhang2023google}.

Scaling the size of the training data and trained models with respect to traditional small-scale speech trainings has posed many challenges beyond engineering efforts and demanding hardware requirements. Among them, a significant challenge was ensuring the convergence of large models, which required adaptations to the learning rate (LR) \citep{whisper,peng24b_interspeech}.
In particular, Whisper \citep{whisper} lowered the peak LR with the increase of the model size. 
Differently, OWSM 3.1 \citep{peng24b_interspeech} introduced a new LR scheduler, driven by the insight that reducing the peak LR would compromise the quality of the trained model \citep{kalra2024why}. The new LR scheduler -- named piecewise LR scheduler -- modifies the warmup phase from a simple linear increase to a two-phase linear warmup while keeping unaltered the decay phase after the LR peak. 
However, this design choice was not motivated, nor was it investigated whether alternative warmup policies could be more effective or how they might impact the final model quality.

\begin{figure*}
    \centering
    \includegraphics[width=1\linewidth]{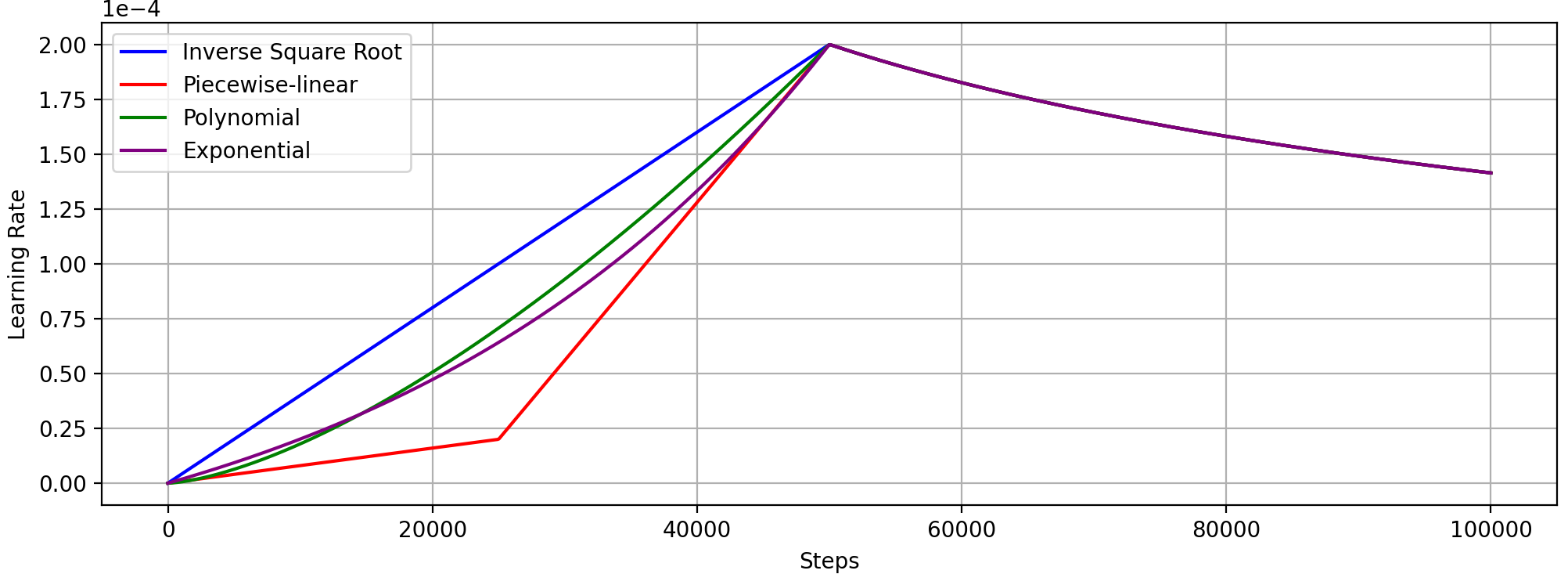}
    \caption{LR schedulers with inverse square root, piecewise-linear, polynomial, and exponential warmup policies.}
    \label{fig:lrs}
\end{figure*}

In this paper, we fill these gaps by studying which factors lead to a more difficult convergence of 
large-scale models and what is the impact of different LR warmup policies on the final performance.
To this aim, we train large-scale S2T Conformer \citep{gulati20_interspeech} models
on more than 150K hours of speech data, exploring alternative warmup methods -- specifically an exponential and a polynomial policy -- operating between the double linear warmup by OWSM and the traditional linear warmup phase of the inverse square root LR scheduler.
Our experiments demonstrate that:

\begin{itemize}
    \item Advanced and more complex variants of the Transformer architecture, such as Conformer and Branchformer \citep{pmlr-v162-peng22a}, widely used in speech processing for their superior
    performance, are more difficult to train due to their deeper layers involving
    additional components (e.g., 
    extra convolutional or linear layers), making them more prone to
    ``exploding gradient'' \citep{279181} issues;
    \item The LR warmup should follow an exponential or sub-exponential function and, while it plays a crucial role in the convergence of the model by ensuring a smooth transition to a good model initialization, it does not significantly affect the final result as long as convergence of the model is achieved.
\end{itemize}

\mmg{To ease future research on the topic, foster reproducibility of our work, and in accordance with the Open Science principles \citep{white2024model}, we release the code, logs, and intermediate checkpoints under the open-source Apache 2.0 license at \url{https://github.com/hlt-mt/FBK-fairseq}.}

\section{Learning Rate Schedulers}

\sara{This section describes the LR schedulers analyzed in this work, starting from the widely adopted inverse square root with linear warmup (\S\ref{subsec:inverse_sqrt}) and piecewise-linear warmup (\S\ref{subsec:piecewise}), to the alternative sub-linear warmup policies, namely polynomial (\S\ref{subsec:polynomial}) and exponential (\S\ref{subsec:exponential}), designed to be as close as possible to the traditional inverse square root LR. All LR schedulers are shown in Figure \ref{fig:lrs}.} 


\subsection{Inverse Square Root Policy}
\label{subsec:inverse_sqrt}

Since the introduction of the Transformer architecture, the LR scheduler has followed an inverse square root policy \citep{attention}. This scheduler has therefore been widely adopted in S2T training settings \citep{inaguma-etal-2020-espnet,wang-etal-2020-fairseq} and entails two phases. Firstly, the LR linearly increases for a predefined number of steps $w$ from 0 to the peak LR $\eta$, where $w$ and $\eta$ are two hyper-parameters whose tuning is critical for the success of the training and the quality of the resulting model \citep{popel-2018}. In this phase, the LR $\eta_i$ at the $i$-th step is $\eta_i = \eta \cdot i / w$. Secondly, after reaching $\eta$, the LR decreases proportionally to the inverse square root of the number of steps, i.e. $\eta_i = \eta \cdot \sqrt{w} / \sqrt{i}$. Overall, the LR $\eta_i$ at the $i$-th step is:

\begin{equation*}
    \eta_i = \eta \cdot \min \left(\frac{i}{w}, \frac{\sqrt{w}}{\sqrt{i}} \right)
\end{equation*}

\noindent where $w$ is set to 50k and $\eta$ to $2e^{-4}$ in this work.

\subsection{Piecewise-linear Warmup}
\label{subsec:piecewise}

\citet{peng24b_interspeech} found that the linear warmup of the standard inverse square root LR scheduler was not suitable for training their large-scale 1B Branchformer model and introduced the piecewise-linear warmup policy. This policy splits the warmup step into two linear phases, introducing an intermediate LR $\eta'$ with a corresponding number of intermediate warmup steps $w'$ as additional hyperparameters. In the first $w'$ steps, the LR linearly increases from 0 to $\eta'$, which is typically set to a much smaller value than $\eta$, and then in the steps between $w'$ and $w$ it increases from $\eta'$ to $\eta$.
As such, in the warmup phase, i.e. at the step $i<w$, the LR  $\eta_i$ is:

\begin{equation*}
    \eta_{i<w} = \max \left( \eta' \cdot \frac{i}{w'}, \eta' +  \frac{(\eta - \eta ')\cdot(i-w')}{w-w'} \right)
\end{equation*}

In this work, we follow \citet{peng24b_interspeech} and set the number of intermediate warmup steps $w'$ to $w/2$ i.e., 25k, and the intermediate LR $w'$ to $\eta/10$.

\subsection{Polynomial Warmup}
\label{subsec:polynomial}

As a first alternative to the piecewise-linear policy, we propose to increase the LR with a polynomial function with respect to the number of steps. The slope of the increase is controlled by a hyper-parameter $\alpha$, according to the formula:

\begin{equation*}
    \eta_{i<w} = \eta \cdot \left( \frac{i}{w} \right)^{\alpha}
\end{equation*}


\sara{We set $\alpha$ to 1.5, and the polynomial warmup function is visualized in Figure \ref{fig:lrs} (green curve).}


\subsection{Exponential Warmup}
\label{subsec:exponential}

As a second alternative, we introduce an exponential policy that, compared to the polynomial one, has a steeper LR increase in the first part of the warmup and a lower LR in the second. Also in this case, the hyper-parameter $\alpha$ controls the smoothness of the function, and the higher the $\alpha$ the smaller the LR in the warmup phase. Specifically, this policy follows the formula:

\begin{equation*}
    \eta_{i<w} = \eta \cdot \frac{e^{\alpha \cdot \frac{i}{w}} - 1}{e^{\alpha} - 1}
\end{equation*}

\sara{Similarly to the polynomial warmup (Section \ref{subsec:polynomial}), we set $\alpha$ to 1.5, and the exponential warmup function is visualized in Figure \ref{fig:lrs} (purple curve).}



\section{Experimental Settings}

To ensure that divergence issues are not due to a particularly challenging setting, we avoided multi-task
trainings, resorting to training S2T models on the automatic speech recognition (ASR) task for two languages (English and Italian). 
\sara{As training data, we use $\sim$150k hours of publicly available speech datasets, which are described in Appendix \ref{app:exp_sett}.}
For validation, we use the English (en) and Italian (it)  dev sets of CommonVoice \citep{ardila-etal-2020-common}.

Our encoder-decoder models have a Transformer decoder and a Conformer encoder preceded by two 1D convolutional layers that downsample the sequence length by a factor of 4. 
\mmg{For the Conformer encoder, we use the implementation by \citet{papi-etal-2024-good} that fixes issues in padding handling.}
\sara{Given the results of preliminary experiments (\S\ref{subsec:preliminary-exp}), we set}
24 encoder layers and 12 decoder layers \sara{for the experiments in \S\ref{subsec:lr-analysis}}. The embeddings have 1024 features, with an FFN hidden dimension of 4096 and 16 attention heads. In total, our models have 878M parameters. 
Further details are provided in Appendix \ref{app:exp_sett}.

\section{Results}
\label{sec:results}

\subsection{Preliminary Experiments}
\label{subsec:preliminary-exp}

In preliminary experiments, we varied the number of encoder and decoder layers to understand when the depth of the network becomes critical -- i.e., the model starts diverging -- with the standard inverse square root LR scheduler.
In this scenario, we observed that the number of encoder layers was the driver of the issue while adding more decoder layers was not. Specifically, models with more than 18 encoder layers were not converging. For instance, models with 18 encoder layers and 6 decoder layers diverge, while models with 12 encoder and 12 decoder layers converge without issues.
This observation, together with the fact that Whisper (which features a Transformer encoder) was trained without the need for adapting the learning rate scheduler, suggests that complex layers featuring many subcomponents, such as Conformer and Branchformer layers, pose convergence issues with deep models. 
\sara{In our Conformer implementation, each subcomponent is wrapped in a residual connection \citep{resnet}, which may indicate a need for additional normalization layers within each encoder block to mitigate potential scaling effects. However, we leave this investigation for future work.}

\subsection{LR Warmup Analysis}
\label{subsec:lr-analysis}
\begin{figure}[h]
    \centering
    \includegraphics[width=1\linewidth]{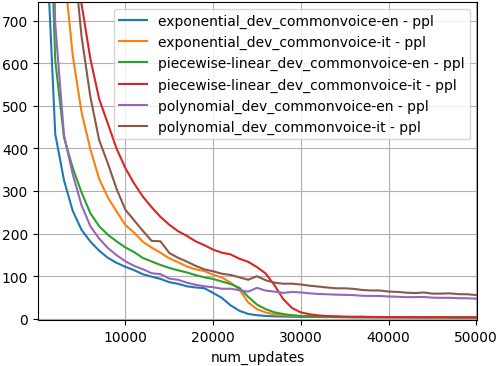}
    \caption{Perplexity on the English and Italian validation sets for the polynomial, piecewise-linear, and exponential policies for the first 50k steps (warmup phase).}
    \label{fig:ppl_dev_1}
\end{figure}

Moving to the comparison of the warmup policies, Figure \ref{fig:ppl_dev_1} shows the resulting learning curves on the validation sets for the two languages, which display the same behaviors, with the only difference that the Italian curves have a higher perplexity at the beginning and decline later than English ones. Similar trends can be observed in the training set, which we report in Appendix \ref{app:ppl_train}.

\paragraph{Model Convergence}  First, we notice that the model convergence is obtained only with the exponential and piecewise-linear policies. The polynomial policy, instead, displays the same pattern as the standard inverse square root policy (which we do not report here) leading the model to a high perplexity that minimally degrades with the progression of the training. This convergence issue can be attributed to an exploding gradient: as we show in Appendix \ref{app:gnorm}, in the polynomial training there are huge spikes in the gradient norm in the range 25k-30k steps and later, where the other policies feature a steep decrease that the polynomial fails to achieve.
The exponential policy, despite a higher LR during the first $\sim$15k steps,
has a slightly lower LR in the 15k-50k range than the polynomial policy.
This minimal difference is 
sufficient to enable model convergence. 
Therefore, we can conclude that the exponential policy closely approaches the highest feasible LR during
the warmup phase without compromising model convergence.

\paragraph{Convergence Speed} Figure \ref{fig:ppl_dev_1} also shows that,
as expected, higher LRs result in lower perplexity during the initial steps.
In both the English and Italian validation sets, the exponential policy -- which features the highest LR in the first $\sim$15k steps -- always displays the lowest perplexity. The polynomial one starts with the highest perplexity
due to its lower LR in the initial steps. However, it later
surpasses the piecewise-linear policy and closes the gap with the exponential one, 
thanks to its higher LR in the later stages, until it ultimately fails to converge.
Interestingly, the learning curves of the two converging policies show a step-like decrease, which is anticipated for the exponential policy ($\sim$20k vs $\sim$23k steps for English and $\sim$22k vs $\sim$26k for Italian) as per its faster convergence.

\begin{table}[]
\small
\begin{tabular}{l|cc|cc|cc|c}
\hline
    \multirow{2}{*}{\textbf{LR}} & \multicolumn{2}{c|}{\textbf{CV}} & \multicolumn{2}{c|}{\textbf{MLS}} & \multicolumn{2}{c|}{\textbf{VP}} & \multirow{2}{*}{\textbf{AVG}} \\
    \cline{2-7}
    & \textit{en}         & \textit{it}        & \textit{en}         & \textit{it}         & \textit{en}        & \textit{it}         &                      \\
\hline
PL  & \textbf{18.4}       & \textbf{13.7}      & \textbf{7.4}        & \textbf{17.4}       & \textbf{8.3}       & \textbf{17.8}       & \textbf{13.8}                 \\

Exp & 19.1       & 14.3      & 7.5        & 17.9       & 8.6       & 18.3       & 14.3      \\
\hline
\end{tabular}
\caption{WER ($\downarrow$), computed using \texttt{jiwer} and the Whisper text normalizer, on the CommonVoice (CV), VoxPopuli (VP), and MLS test sets of the 170k-steps checkpoints obtained with the LR scheduler with piecewise-linear (PL) and exponential (Exp) warm up.}
\label{tab:wer}
\end{table}

\begin{figure}[h]
    \centering
    \includegraphics[width=1\linewidth]{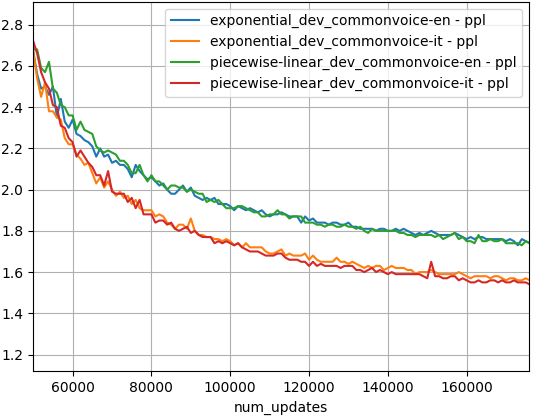}
    \caption{Perplexity on the English and Italian validation sets for the piecewise-linear and exponential policies for the steps after the warmup phase (50k-170k).}
    \label{fig:ppl_dev_2}
\end{figure}

\paragraph{Effect on the Resulting Model} Lastly, we explore whether the faster initial convergence of the exponential policy results in a better model at the end of the training compared to that obtained with the piecewise-linear policy. Figure \ref{fig:ppl_dev_2} shows the learning curve after the first 50k steps, up to the end of the whole pass over the training set (i.e., 
the first training epoch at step 170k). The learning curves of the piecewise-linear scheduler not only reach the perplexity of those of the exponential policy but the English one also becomes slightly better. The same trend is observed in the training data (see Figure \ref{fig:ppl_train_2} in Appendix \ref{app:ppl_train}), in which the English data is more than 80\%. The WER on test sets for the checkpoint at the 170k step also testifies to a slight superiority of the model obtained using the piecewise-linear policy on both languages, as shown in Table \ref{tab:wer}. We can conclude that a faster convergence in the early stages of the training does not imply a better resulting model and that the warmup policy of the LR scheduler is critical to ensure the convergence of the model, but, once that is achieved, its role in the model quality is limited.

\section{Conclusions}

In this study, we analyzed one of the key challenges -- beyond engineering, data curation, and hardware efforts -- associated with training large-scale S2T models i.e., the role of the LR scheduler and, in particular, of its warmup strategy in model convergence and final performance.
To this aim, we compared the standard linear warmup and the piecewise-linear warmup strategies with two policies -- polynomial and exponential -- aimed at finding the highest possible LR in the warmup phase that does not lead to convergence issues. Through experiments on large-scale ASR trainings of a $\sim$900M parameters Conformer model, we demonstrated that while the LR warmup phase is crucial for stabilizing convergence, it has a minimal impact on final model performance and that the LR warmup phase should follow an exponential or sub-exponential rise to ensure model convergence.

\section*{Acknowledgments}

 \lb{This paper has received funding from the PNRR project FAIR - Future AI Research (PE00000013),  under the NRRP MUR program funded by the NextGenerationEU, and from the European Union’s Horizon research and innovation programme under grant agreement No 101135798, project Meetween (My Personal AI Mediator for Virtual MEETings BetWEEN People).
We acknowledge CINECA for the availability of high-performance computing resources and support.}

\section*{Limitations}

\paragraph{Effect of Multilingualism and Multi-task} In this work, we decided to experiment with a single task and two languages in the training, even though the amount of training data we used was comparable to that used in other works to train S2T models on multiple tasks and more than 100 languages (e.g., OWSM uses 180k hours of data against our 150k hours). Although there is no reason to posit that a different setting may lead to different conclusions since the behaviors we observed were similar to those of OWSM, future works should validate that our findings extend to these scenarios.

\paragraph{Multiple Runs} While performing multiple runs for each setting would provide stronger insights into the possible statistical significance of the observed differences, this would require extensive computational costs that go beyond our budget.

\paragraph{Tuning $\alpha$} Although by tuning $\alpha$ we could, for instance, obtain a converging model even with the polynomial policy, this was not the focus of our work. In this paper, we attempted to understand the role of different LR schedulers on the resulting model and what could be achieved by using different LR warmup policies. Since two extreme solutions -- the piecewise-linear policy with a relatively low LR and the exponential policy with the highest feasible LR -- do not show evident differences, finding other values of $\alpha$ or other policies leading to similar results would not have added much to our discussion. Also, as noted above, each run is computationally demanding, limiting our ability to explore the space of the possible values.

\bibliography{custom}

\newpage
\appendix

\section{Training Settings}
\label{app:exp_sett}

\sara{We train the models on $\sim$150k hours of speech datasets, namely the train section of CommonVoice \citep{ardila-etal-2020-common}, CoVoST2 \citep{wang21s_interspeech}, FLEURS \citep{10023141}, LibriLight \citep{librilight}, MLS \citep{pratap20_interspeech}, VoxPopuli \citep{wang-etal-2021-voxpopuli}, and YouTube-Commons \citep{huggingfacePleIAsYouTubeCommonsDatasets}. 
\mmg{When the transcript was not available for a given dataset, we used the automatic transcripts of MOSEL v1.0 \citep{mosel}. As YouTube-Commons transcripts are not available in MOSEL v1.0\footnote{They have been added in v2.0.}, we used the transcript provided for the training of FAMA \citep{papi2025fama}. Our training data is exactly the same used for FAMA and is available at \url{https://huggingface.co/datasets/FBK-MT/fama-data}.}
The textual data is used to build the vocabulary with 16,000 SentencePiece unigrams \citep{kudo-2018-subword}.}

We optimize our models using the Adam optimizer with betas (0.9, 0.98).
The training loss is the linear combination of the label-smoothed cross-entropy \citep{szegedy2016rethinking} on the decoder output and two CTC \citep{Graves2006ConnectionistTC} losses, one at the 16th encoder layer and one on top of the encoder \citep{9003774,yan-etal-2023-ctc}. We also experimented with removing the auxiliary CTC losses, to ensure that they were not the driver of divergence issues and, indeed, their removal did not change anything in terms of whether a model converges or not. We clip the gradient norm at 10.0 and use 0.001 weight decay. We trained the models on 16 A100 GPUs (64GB VRAM) for 1 epoch with at most 55 seconds of data in each mini-batch and 5 gradient accumulation steps, resulting in 176,208 batches to complete an epoch. One run in this setting lasts 6 days.

\section{Perplexity on the Training Set}
\label{app:ppl_train}

\begin{figure}[h]
    \centering
    \includegraphics[width=0.95\linewidth]{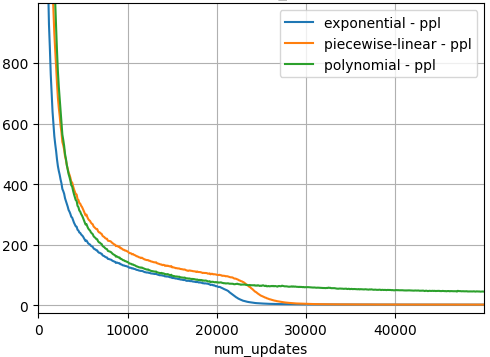}
    \caption{Perplexity on the training set for the polynomial, piecewise-linear, and exponential warmup policies for the first 50k steps (warmup phase).}
    \label{fig:ppl_train_1}
\end{figure}

Figure \ref{fig:ppl_train_1} shows the perplexity (PPL) of the different warmup policies on the training set for the first part of the training. Compared to Figure \ref{fig:ppl_dev_1} presenting the PPL obtained on the validation set, the training curves show similar behaviors, with the polynomial warmup not converging, and the piecewise-linear and exponential leading to, respectively, slower and faster convergence.

\begin{figure}[h]
    \centering
    \includegraphics[width=0.95\linewidth]{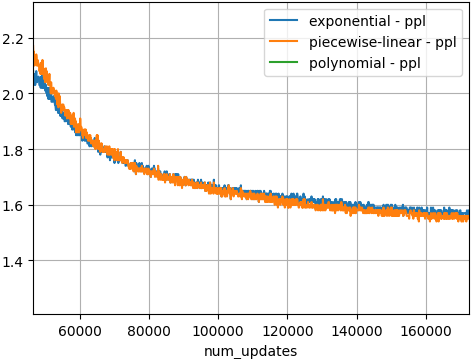}
    \caption{Perplexity on the training set for the piecewise-linear and exponential warmup policies for the steps after the warmup phase (50k-170k).}
    \label{fig:ppl_train_2}
\end{figure}

\begin{figure*}
    \centering
    \includegraphics[width=0.95\linewidth]{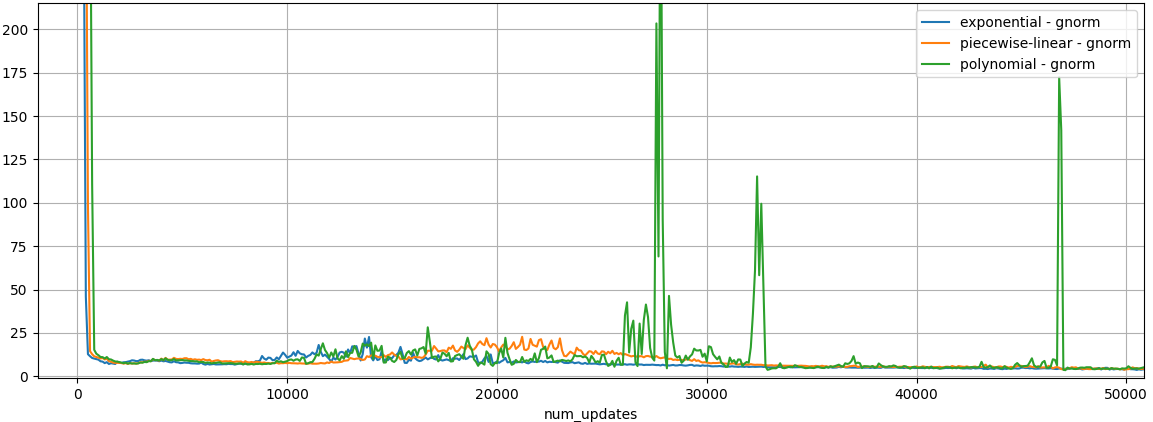}
    \caption{Gradient norm comparison across the piecewise-linear, polynomial, and exponential warmup policies.}
    \label{fig:gnorm}
\end{figure*}

Looking at Figure \ref{fig:ppl_train_2} that isolates the PPL behavior after the first 50k steps, we notice that, again, the piecewise-linear and exponential warmup exhibit similar trends to those reported for the validation set in Figure \ref{fig:ppl_dev_2}: the curves are very close, with the piecewise-linear, initially above the exponential, becoming slightly below the exponential in the long run.
This reconfirms the results discussed in Section \ref{sec:results}, where we highlighted the convergence issues of the polynomial function, which is actually reflected in the training set, and the slower but slightly better convergence of the piecewise-linear warmup against the exponential one.

\section{Gnorm Analysis}
\label{app:gnorm}

Figure \ref{fig:gnorm} reports the gradient norm in the warmup phase for the different policies (exponential, polynomial, and piecewise-linear). Except for the initial steps, the gradient norm for the policies leading to convergence always remains low ($<$25). For the polynomial warmup, instead, there are huge spikes beyond 100 and even 200 after 25k steps. These explosions of the gradient norm have also been observed in all the runs with the inverse square root LR scheduler that did not converge in our preliminary experiments. We can conclude that huge spikes in the gradient norm can be used to detect non-converging trainings.

Analyzing the gradient norm of the exponential and piecewise-linear policies, we observe that the gradient norm is higher at the beginning (8k-15k steps) for the exponential policy, which displays faster convergence in this phase. On the opposite, the gradient norm of the piecewise-linear policy is higher in the 15k-30k steps range, in which closes the initial gap with the exponential policy.

\end{document}